\documentclass[11pt,a4paper]{article}

\usepackage[margin=1in]{geometry}
\usepackage{tikz}
\usetikzlibrary{positioning, arrows.meta, fit, calc, backgrounds}
\usepackage{booktabs}
\usepackage{algorithm}
\usepackage{algpseudocode}
\usepackage{subcaption}
\usepackage{pgfplots}
\pgfplotsset{compat=1.18}
\usepgfplotslibrary{groupplots}
\usepackage{amsmath,amssymb}
\usepackage{hyperref}
\usepackage[numbers]{natbib}
\usepackage{authblk}

\hypersetup{
  colorlinks=true,
  linkcolor=blue!70!black,
  citecolor=blue!70!black,
  urlcolor=blue!70!black
}

\title{Personality Requires Struggle: Three Regimes of the Baldwin Effect in Neuroevolved Chess Agents}

\author[1]{Diego Armando Resendez Prado}
\affil[1]{Independent Researcher \\ \texttt{diego.resendez@zero-oneit.com}}

\date{\today}

\begin{document}

\maketitle

\begin{abstract}
Can lifetime learning expand behavioral diversity over evolutionary time, rather than collapsing it?
Prior theory predicts that plasticity reduces variance by buffering organisms against environmental noise.
We test this in a competitive domain: chess agents with eight NEAT-evolved neural modules, Hebbian within-game plasticity, and a desirability-domain signal chain with imagination.
Across 10~seeds per Hebbian condition, a \emph{variance crossover} emerges: Hebbian ON starts with lower cross-seed variance than OFF, then surpasses it at generation~34.
The crossover trend is monotonic ($\rho = 0.91$, $p < 10^{-6}$): plasticity's effect on behavioral variance reverses over evolutionary time, initially compressing diversity (consistent with prior predictions) then expanding it as evolved Perception differences are amplified through imagination---a feedback loop that mutation alone cannot sustain.

The result is structured behavioral divergence: evolved agents select different moves on the same positions (62\% disagreement), develop distinct opening repertoires, piece preferences, and game lengths.
These are not different sampling policies---they are reproducible behavioral signatures (ICC $> 0.8$) with interpretable signal chain configurations.
Three regimes appear depending on opponent type: exploration (Hebbian ON, heterogeneous opponent), lottery (Hebbian OFF, elitism lock-in), and transparent (same-model opponent, brain self-erasure).
The transparent regime generates a falsifiable prediction: self-play systems may systematically suppress behavioral diversity by eliminating the heterogeneity that personality requires.

\end{abstract}

\section{Introduction}

Does lifetime learning help or hinder the evolution of diverse behaviors?
Theory predicts that plasticity should \emph{reduce} behavioral variance by buffering organisms against environmental noise~\citep{paenke2009baldwin}, accelerating convergence toward a shared optimum.
But this prediction rests on toy domains---binary strings, simple function landscapes---where the fitness landscape has a single basin of attraction.
In a competitive domain with a complex fitness landscape, plasticity might instead open new regions of the landscape, producing \emph{more} diverse organisms over time.
We test this question empirically, using chess as a domain where both performance and behavioral diversity can be measured.

Prior work~\citep{resendez2026ailed} introduced a hand-crafted signal chain that reshapes a chess predictor's move distribution, producing engines with measurable play style differences.
We replace the hand-crafted layer with an evolved one: \emph{what happens if you let evolution design the cognitive architecture, and if the evolving population can also learn within its lifetime?}
We define \emph{behavioral divergence} as systematic, seed-specific deviation from the cartridge's default move distribution, measured by the intraclass correlation coefficient (ICC) of agreement trajectories across seeds.
An ICC above 0.75 indicates that between-seed variance dominates within-seed variance---seeds maintain distinct, reproducible behavioral signatures rather than fluctuating randomly.
Throughout this paper, we use ``personality'' as shorthand for this measurable behavioral divergence (ICC $= 0.81$ for Hebbian ON, $0.92$ for OFF; both qualify as strong consistency).

This second question has a long history in evolutionary biology.
In 1896, James Mark Baldwin proposed that organisms capable of learning within their lifetimes could gain evolutionary advantages, not because learned traits are inherited directly, but because learning ability itself is heritable and buffers organisms against environmental variation~\citep{baldwin1896}.
Hinton and Nowlan~\citep{hinton1987} demonstrated this computationally: in a needle-in-a-haystack fitness landscape, a population with lifetime learning found the optimum far faster than one relying on mutation alone.
Most subsequent computational studies have used toy domains: binary strings, simple function optimization.
How the Baldwin Effect behaves in a multi-module cognitive architecture remains untested.
We do not claim to have fully demonstrated the Baldwin Effect (which would require tracking genetic assimilation across hundreds of generations); rather, we identify three regimes of Baldwin-compatible dynamics and characterize how plasticity interacts with evolution to shape behavioral diversity.

We apply this idea to chess.
AILED-Brain replaces the hand-crafted psyche of~\citet{resendez2026ailed} with eight NEAT-evolved neural modules (Perception, Memory, Affect, Attention, Dynamics, Personality, Integration, and Learning) whose outputs drive a desirability-domain signal chain.
The chain includes an imagination step---1-ply subjective lookahead through the brain's own evolved Perception---and per-piece preference weights.
Within each game, Hebbian plasticity adjusts connection weights based on board outcomes.
After the game, NEAT selection, crossover, and mutation evolve the population.
This two-speed system sets up the Baldwin Effect.

The architecture borrows labels from human cognition---perception, memory, affect, attention---but evolution found different uses for them.
A key enabler is the underlying move predictor: the AILED Engine v26.3.0, a 43M-parameter encoder-decoder transformer that exposes win/draw/loss estimates and piece-level attention alongside move probabilities.
We are not aware of another chess engine that exposes this level of perceptual detail to an external cognitive layer.
By testing the brain on this expressive cartridge against a heterogeneous opponent (Maia2 at 1100~Elo) with 10~seeds per Hebbian condition, and comparing against same-model matchups at various Elo levels, we discover three qualitatively different regimes of evolved organism.
Non-trivial signal chain preferences appear only when the cartridge and opponent are different engines.
Same-model matchups produce brains that erase themselves.

\paragraph{Contributions.}
(1)~Under the conditions tested, plasticity expands behavioral diversity over evolutionary time: a variance crossover where Hebbian ON starts with \emph{lower} cross-seed variance than OFF, then surpasses it, increasing monotonically across 50~generations (Spearman $\rho = 0.91$, $p < 10^{-6}$).
This extends prior theory~\citep{paenke2009baldwin}, which predicted variance reduction, by showing the effect reverses at longer timescales.
(2)~Three regimes of Baldwin-compatible dynamics---exploration, lottery, and transparent---replicated with 10~seeds per condition and validated under alternative fitness weights.
(3)~Structured behavioral divergence: evolved agents select different moves on the same board (62\% disagreement), develop distinct opening repertoires, piece preferences, and game lengths, with ICC $> 0.8$ across generations.
(4)~A falsifiable prediction: same-model opponents produce self-erasing brains regardless of skill gap, suggesting self-play systems may systematically suppress behavioral diversity.
(5)~A test of evolutionary dynamics with lifetime learning in a competitive domain, moving past toy-problem demonstrations.

\section{Background and Related Work}

\subsection{The Baldwin Effect}

The Baldwin Effect~\citep{baldwin1896} describes how organisms capable of lifetime learning can gain evolutionary advantages.
The core mechanism is indirect: learning does not alter the genome, but it changes which genomes are selected.
An organism that can \emph{learn} a beneficial behavior survives and reproduces even if its genome does not encode that behavior directly.
Over generations, genetic variants that approximate the learned behavior gain a selective advantage.

Hinton and Nowlan~\citep{hinton1987} demonstrated this computationally with a binary string optimization task.
Ackley and Littman~\citep{ackley1991} extended the result to continuous fitness landscapes.
Turney~\citep{turney1996baldwin} clarified common misconceptions, showing that the Baldwin Effect can both accelerate and decelerate evolution depending on environmental stability.
Paenke et al.~\citep{paenke2009baldwin} formalized conditions under which plasticity helps versus hinders directional selection, predicting that plasticity generally reduces phenotypic variance by buffering organisms against environmental noise.
Soltoggio et al.~\citep{soltoggio2018neuromodulation} surveyed evolved plastic neural networks, noting that neuromodulation and Hebbian learning are the most common mechanisms for lifetime adaptation.

Prior computational demonstrations, however, stay in toy domains.
We test the Baldwin Effect in a competitive game where both performance and personality can be measured.

\subsection{Neuroevolution and NEAT}

NEAT~\citep{stanley2002neat} evolves both the topology and weights of neural networks, starting from minimal structures and complexifying through mutation operators that add nodes and connections.
NEAT has been applied to game AI~\citep{stanley2005rtNEAT, risi2017neuroevolution}, but typically evolves a single monolithic network rather than a modular architecture with specialized components.
Clune et al.~\citep{clune2013modularity} showed that modularity emerges in evolved networks when the environment contains separable subproblems.
Our architecture imposes modular structure a priori---eight modules with defined roles---while allowing NEAT to evolve each module's internal topology independently.

\subsection{Human-Like Chess Engines}

The dominant paradigm in chess AI optimizes for playing strength~\citep{silver2018alphazero}.
Maia~\citep{mcilroyyoung2020maia} and Maia2~\citep{mcilroyyoung2022maia2} broke from this paradigm by training on human games at specific skill levels, producing engines that predict human moves rather than optimal ones.
Prior work~\citep{resendez2026ailed} introduced emotional state modeling to a prediction-based engine, showing that a hand-crafted psyche layer produces measurable play style differences.
AILED-Brain takes the next step: rather than designing the emotional layer, we evolve it and study what kinds of organisms the evolutionary process produces under different conditions.

\section{Architecture}

\subsection{Pool Architecture}

\begin{figure}[t]
\centering
\resizebox{0.8\columnwidth}{!}{%
\begin{tikzpicture}[
    pool/.style={draw, rounded corners, minimum width=6.8cm, minimum height=0.7cm, font=\small\sffamily, fill=gray!10},
    module/.style={draw, rounded corners, minimum width=1.8cm, minimum height=0.7cm, font=\scriptsize\sffamily},
    phase/.style={draw, rounded corners, minimum width=2.4cm, minimum height=0.7cm, font=\small\sffamily},
    chain/.style={draw, rounded corners, minimum width=2.4cm, minimum height=0.7cm, font=\small\sffamily, fill=orange!8},
    arrow/.style={-{Stealth[length=2.5mm]}, thick},
    label/.style={font=\tiny, text=gray},
]
    \node[pool] (pool) {Signal Pool: sensors (20) + game ctx (8) + WDL (3) + dist shape (4)};

    \node[module, fill=blue!15, below left=0.8cm and 2.8cm of pool] (perc) {Perception};
    \node[module, fill=green!15, right=0.2cm of perc] (mem) {Memory};
    \node[module, fill=red!15, right=0.2cm of mem] (aff) {Affect};
    \node[module, fill=yellow!15, right=0.2cm of aff] (att) {Attention};
    \node[module, fill=purple!12, right=0.2cm of att] (dyn) {Dynamics};

    \node[label, left=0.1cm of perc] {\textbf{Phase 1}};

    \draw[arrow, gray!60] (pool.south) -- ++(0,-0.3) -| (perc.north);
    \draw[arrow, gray!60] (pool.south) -- ++(0,-0.3) -| (mem.north);
    \draw[arrow, gray!60] (pool.south) -- ++(0,-0.3) -| (aff.north);
    \draw[arrow, gray!60] (pool.south) -- ++(0,-0.3) -| (att.north);
    \draw[arrow, gray!60] (pool.south) -- ++(0,-0.3) -| (dyn.north);

    \node[phase, fill=purple!20, below=0.8cm of aff] (pers) {Personality};
    \node[label, left=0.6cm of pers] {\textbf{Phase 2}};

    \draw[arrow] (perc.south) -- ++(0,-0.25) -| (pers.north west);
    \draw[arrow] (mem.south) -- ++(0,-0.25) -| ([xshift=-0.4cm]pers.north);
    \draw[arrow] (aff.south) -- (pers.north);
    \draw[arrow] (att.south) -- ++(0,-0.25) -| ([xshift=0.4cm]pers.north);
    \draw[arrow] (dyn.south) -- ++(0,-0.25) -| (pers.north east);

    \node[phase, fill=teal!15, below=0.6cm of pers] (integ) {Integration};
    \node[label, left=0.6cm of integ] {\textbf{Phase 3}};
    \draw[arrow] (pers) -- (integ);

    \node[chain, below=0.6cm of integ] (sc) {Signal Chain + Piece EQ};
    \draw[arrow] (integ) -- node[right, font=\tiny] {10 params + 6 wt} (sc);

    \node[chain, below=0.4cm of sc] (imag) {Imagination (1-ply)};
    \draw[arrow] (sc) -- (imag);

    \draw[arrow, dashed, blue!50] (perc.west) -- ++(-0.4,0) |- (imag.west);

    \node[chain, below=0.4cm of imag] (move) {\textbf{Move}};
    \draw[arrow] (imag) -- (move);

\end{tikzpicture}
}
\caption{Pool architecture. Phase~1: five modules read independently from a shared signal pool (board sensors, game context, cartridge WDL, distribution shape). Phase~2: Personality reads all Phase~1 outputs. Phase~3: Integration produces signal chain parameters. The signal chain reshapes the cartridge's move distribution; imagination evaluates top candidates through 1-ply lookahead using Perception (dashed). Each module is a NEAT network.}
\label{fig:pipeline}
\end{figure}
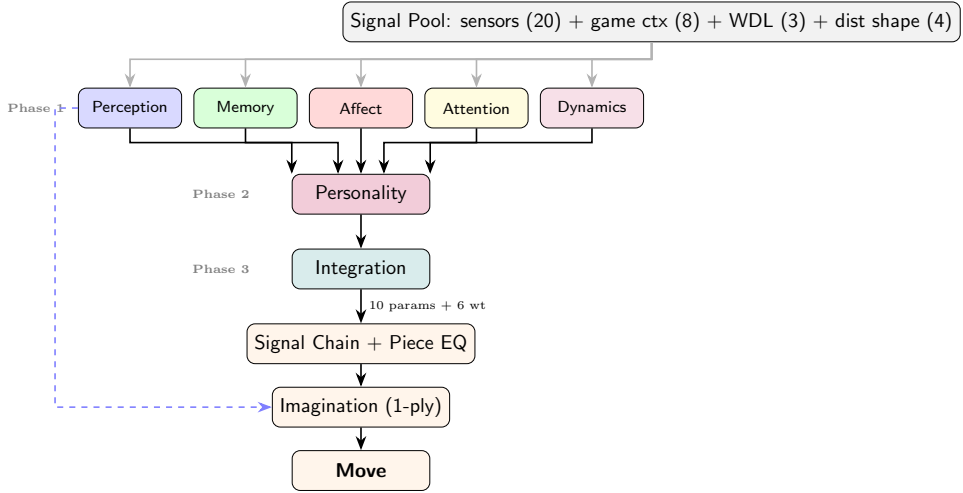

AILED-Brain processes each position through a three-phase pool architecture (Figure~\ref{fig:pipeline}).
Each module is a small NEAT network with typed input and output nodes.
In Phase~1, five modules (Perception, Memory, Affect, Attention, Dynamics) read independently from a shared signal pool---board sensors, game context, cartridge WDL, and distribution shape.
In Phase~2, Personality reads all Phase~1 outputs plus the pool.
In Phase~3, an Integration module combines everything into signal chain parameters.

\paragraph{Board Sensors.}
Twenty normalized features are extracted from the board position: material balance, king exposure, pawn structure (passed, isolated, doubled), center control, piece mobility, rook activity, bishop pair bonus, knight outposts, piece coordination, king pawn shield, open files, and game phase.
All values are scaled to $[0, 1]$.

\paragraph{Module Roles.}
Table~\ref{tab:modules} summarizes each module's inputs, outputs, and execution phase.
Phase~1 modules run in parallel---each reads a different slice of the signal pool without seeing the others' outputs.
Personality (Phase~2) then reads all five Phase~1 outputs, acting as the brain's preference layer.
Integration (Phase~3) combines everything into 16 values mapped to signal chain parameters and piece weights.

\begin{table}[t]
\caption{Cognitive modules and execution phases.}
\label{tab:modules}
\centering
\footnotesize
\begin{tabular}{@{}rlll@{}}
\toprule
Ph & Module & Dims & Role \\
\midrule
1 & Perception & 20 $\to$ 8 & Board feature compression \\
1 & Memory & 8 $\to$ 4 & Game trajectory context \\
1 & Affect & 8 $\to$ 5 & Emotional response (WDL) \\
1 & Attention & 16 $\to$ 4 & Selective focus (pieces) \\
1 & Dynamics & 6 $\to$ 2 & Confidence calibration \\
\midrule
2 & Personality & 27 $\to$ 8 & Preference shaping \\
3 & Integration & 31 $\to$ 16 & Param synthesis \\
-- & Learning & 4 $\to$ 6 & Post-game adaptation \\
\bottomrule
\end{tabular}
\end{table}

\begin{figure*}[t]
\centering
\resizebox{\textwidth}{!}{%
\begin{tikzpicture}[
    brainbox/.style={draw, rounded corners, minimum width=2.6cm, minimum height=0.65cm, font=\small\sffamily, thick},
    ailedbox/.style={draw, rounded corners, minimum width=2.6cm, minimum height=0.65cm, font=\small\sffamily, thick},
    mapline/.style={-{Stealth[length=2mm]}, dashed, gray!60, thick},
    phaselabel/.style={font=\scriptsize\sffamily, text=gray},
    desc/.style={font=\scriptsize, text=black!70, align=left},
    heading/.style={font=\large\sffamily\bfseries},
]

\node[heading] at (-4.5, 7.2) {Human Brain};

\node[brainbox, fill=gray!15] (eyes) at (-4.5, 6.2) {Eyes / Senses};

\node[brainbox, fill=blue!15] (v1) at (-6.8, 4.5) {Visual Cortex};
\node[desc, right] at (-5.4, 4.5) {\strut V1--V4};
\node[brainbox, fill=green!15] (hippo) at (-6.8, 3.5) {Hippocampus};
\node[desc, right] at (-5.4, 3.5) {\strut episodic memory};
\node[brainbox, fill=red!15] (amyg) at (-6.8, 2.5) {Amygdala};
\node[desc, right] at (-5.4, 2.5) {\strut emotion / threat};
\node[brainbox, fill=yellow!15] (parietal) at (-6.8, 1.5) {Parietal Cortex};
\node[desc, right] at (-5.4, 1.5) {\strut salience / focus};
\node[brainbox, fill=purple!12] (cerebellum) at (-6.8, 0.5) {Cerebellum};
\node[desc, right] at (-5.4, 0.5) {\strut calibration};

\node[phaselabel] at (-8.8, 2.5) {parallel};

\node[brainbox, fill=purple!20] (pfc) at (-4.5, -0.8) {Prefrontal Cortex};
\node[desc, right] at (-3.0, -0.8) {\strut personality};

\node[brainbox, fill=teal!15] (bg) at (-4.5, -2.0) {Basal Ganglia};
\node[desc, right] at (-3.0, -2.0) {\strut action selection};

\node[brainbox, fill=orange!15] (motor) at (-4.5, -3.2) {Motor Cortex};
\node[desc, right] at (-3.0, -3.2) {\strut execute action};

\node[brainbox, fill=blue!8, draw=blue!40] (imagery) at (-4.5, -4.6) {Mental Imagery};
\node[desc, right] at (-3.0, -4.6) {\strut reuses visual cortex};

\draw[-{Stealth}, gray!60] (eyes) -- ++(0,-0.8) -| (v1.north);
\draw[-{Stealth}, gray!60] (eyes) -- ++(0,-0.8) -| (hippo.north);
\draw[-{Stealth}, gray!60] (eyes) -- ++(0,-0.8) -| (amyg.north);
\draw[-{Stealth}, gray!60] (eyes) -- ++(0,-0.8) -| (parietal.north);
\draw[-{Stealth}, gray!60] (eyes) -- ++(0,-0.8) -| (cerebellum.north);
\draw[-{Stealth}] (v1.south) -- ++(0,-0.3) -| (pfc.north west);
\draw[-{Stealth}] (hippo.south) -- ++(0,-0.3) -| ([xshift=-0.5cm]pfc.north);
\draw[-{Stealth}] (amyg.south) -- ++(0,-0.3) -| (pfc.north);
\draw[-{Stealth}] (parietal.south) -- ++(0,-0.3) -| ([xshift=0.5cm]pfc.north);
\draw[-{Stealth}] (cerebellum.south) -- ++(0,-0.3) -| (pfc.north east);
\draw[-{Stealth}] (pfc) -- (bg);
\draw[-{Stealth}] (bg) -- (motor);
\draw[-{Stealth}] (motor) -- (imagery);
\draw[-{Stealth}, dashed, blue!50] (v1.west) -- ++(-0.3,0) |- (imagery.west);

\node[heading] at (4.5, 7.2) {AILED-Brain};

\node[ailedbox, fill=gray!15, minimum width=3.0cm] (pool) at (4.5, 6.2) {Signal Pool};

\node[ailedbox, fill=blue!15] (perc) at (2.2, 4.5) {Perception};
\node[desc, right] at (3.6, 4.5) {\strut 20$\to$8};
\node[ailedbox, fill=green!15] (mem) at (2.2, 3.5) {Memory};
\node[desc, right] at (3.6, 3.5) {\strut 8$\to$4};
\node[ailedbox, fill=red!15] (aff) at (2.2, 2.5) {Affect};
\node[desc, right] at (3.6, 2.5) {\strut 8$\to$5};
\node[ailedbox, fill=yellow!15] (att) at (2.2, 1.5) {Attention};
\node[desc, right] at (3.6, 1.5) {\strut 16$\to$4};
\node[ailedbox, fill=purple!12] (dyn) at (2.2, 0.5) {Dynamics};
\node[desc, right] at (3.6, 0.5) {\strut 6$\to$2};

\node[phaselabel] at (0.2, 2.5) {Phase 1};

\node[ailedbox, fill=purple!20] (pers) at (4.5, -0.8) {Personality};
\node[desc, right] at (6.0, -0.8) {\strut 27$\to$8};
\node[phaselabel] at (0.2, -0.8) {Phase 2};

\node[ailedbox, fill=teal!15] (integ) at (4.5, -2.0) {Integration};
\node[desc, right] at (6.0, -2.0) {\strut 31$\to$16};
\node[phaselabel] at (0.2, -2.0) {Phase 3};

\node[ailedbox, fill=orange!15] (sc) at (4.5, -3.2) {Signal Chain};
\node[desc, right] at (6.0, -3.2) {\strut 10 params + 6 wt};

\node[ailedbox, fill=blue!8, draw=blue!40] (imag) at (4.5, -4.6) {Imagination};
\node[desc, right] at (6.0, -4.6) {\strut 1-ply via Perception};

\draw[-{Stealth}, gray!60] (pool) -- ++(0,-0.8) -| (perc.north);
\draw[-{Stealth}, gray!60] (pool) -- ++(0,-0.8) -| (mem.north);
\draw[-{Stealth}, gray!60] (pool) -- ++(0,-0.8) -| (aff.north);
\draw[-{Stealth}, gray!60] (pool) -- ++(0,-0.8) -| (att.north);
\draw[-{Stealth}, gray!60] (pool) -- ++(0,-0.8) -| (dyn.north);
\draw[-{Stealth}] (perc.south) -- ++(0,-0.3) -| (pers.north west);
\draw[-{Stealth}] (mem.south) -- ++(0,-0.3) -| ([xshift=-0.5cm]pers.north);
\draw[-{Stealth}] (aff.south) -- ++(0,-0.3) -| (pers.north);
\draw[-{Stealth}] (att.south) -- ++(0,-0.3) -| ([xshift=0.5cm]pers.north);
\draw[-{Stealth}] (dyn.south) -- ++(0,-0.3) -| (pers.north east);
\draw[-{Stealth}] (pers) -- (integ);
\draw[-{Stealth}] (integ) -- (sc);
\draw[-{Stealth}] (sc) -- (imag);
\draw[-{Stealth}, dashed, blue!50] (perc.west) -- ++(0.0,0) -- ++(-1.2,0) |- (imag.west);

\draw[mapline] (v1.east) -- (perc.west);
\draw[mapline] (hippo.east) -- (mem.west);
\draw[mapline] (amyg.east) -- (aff.west);
\draw[mapline] (parietal.east) -- (att.west);
\draw[mapline] (cerebellum.east) -- (dyn.west);
\draw[mapline] (pfc.east) -- (pers.west);
\draw[mapline] (bg.east) -- (integ.west);
\draw[mapline] (motor.east) -- (sc.west);
\draw[mapline] (imagery.east) -- (imag.west);

\node[font=\scriptsize\sffamily, text=gray, align=center] at (-4.5, -5.8) {Synaptic plasticity (LTP/LTD)\\$\updownarrow$\\Natural selection (DNA)};
\node[font=\scriptsize\sffamily, text=gray, align=center] at (4.5, -5.8) {Hebbian learning ($\Delta w = \eta \cdot a \cdot r$)\\$\updownarrow$\\NEAT evolution (genome)};

\end{tikzpicture}
}
\caption{Neuroscience inspiration for the AILED-Brain architecture. Left: human brain regions. Right: corresponding AILED-Brain modules. Dashed gray arrows show the functional mapping; color coding indicates analogous roles. Both systems feature parallel sensory processing, preference integration, and action selection. The key parallel: imagination reuses Perception (blue dashed) the same way mental imagery reuses visual cortex. Bottom: both operate on two timescales---within-lifetime plasticity and cross-generation evolution.}
\label{fig:neuro}
\end{figure*}
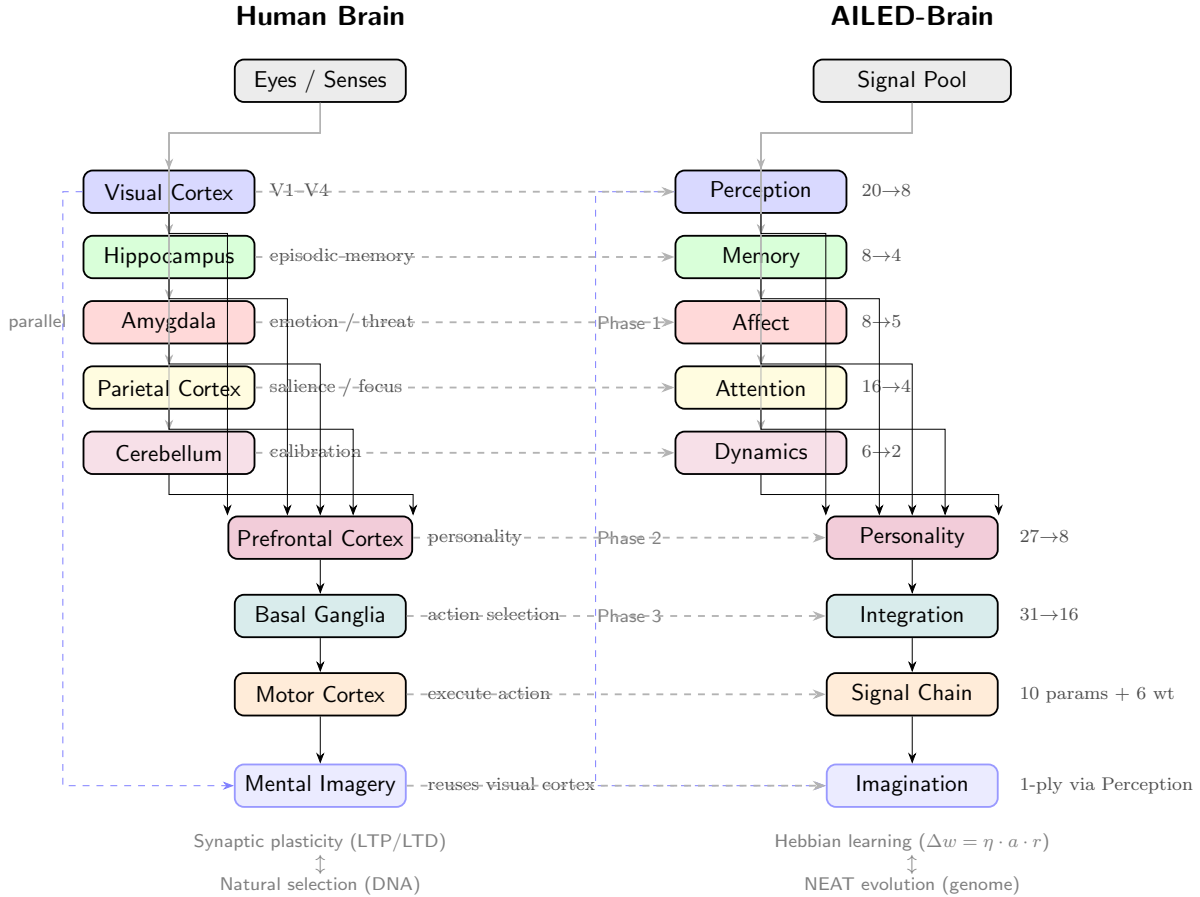

\paragraph{Neuroscience inspiration.}
The modular architecture draws on the functional organization of the human brain (Figure~\ref{fig:neuro}).
These labels are aspirational---evolution is free to repurpose any module, and the results show that it does.
One parallel deserves emphasis: imagination reuses Perception the same way the human brain reuses visual cortex for mental imagery~\citep{kosslyn2001neural}.
Because each seed evolves different Perception weights, the same future position ``feels'' different to different brains---the mechanism that drives seed-specific behavioral divergence.
The two-timescale learning system (Hebbian within-game, NEAT across generations) maps to synaptic plasticity (LTP/LTD) and phylogenetic adaptation, respectively---the same gap that the Baldwin Effect bridges in biological evolution.

\subsection{Desirability-Domain Signal Chain}

The Integration module outputs ten continuous parameters that drive an audio-inspired signal chain~\citep{resendez2026ailed}.
The chain operates in the \emph{desirability} domain rather than directly on move probabilities: the brain expresses what it wants independently of what the predictor predicts.
Given raw logits $\ell_i$ for legal move $i$, the chain computes:
\begin{equation}
    d_i = \begin{cases} \ell_i^\alpha & \text{if } p_i > \tau \text{ (gate)} \\ -\infty & \text{otherwise} \end{cases}
    \label{eq:desirability}
\end{equation}
where $\alpha \in [0.3, 2.5]$ is dynamics compression and $\tau \in [0, 0.3]$ is the probability gate threshold ($p_i = \text{softmax}(\ell)_i$).
The surviving desirabilities are then modified by five-band equalization ($d_i \mathrel{+}= \ln g_b \cdot s$ for each band $b$ with gain $g_b \in [0.1, 3.0]$, sensitivity $s = 0.4$) and per-piece weights ($d_i \mathrel{+}= \ln w_t \cdot s$ for piece type $t$).
Finally, $\mathbf{p} = \text{softmax}(\mathbf{d} / T)$ with temperature $T = 1 + 0.5 \cdot \Delta T$ ($\Delta T \in [-1, 1]$), followed by saturation clamping and exploration bonus.

\paragraph{Imagination.}
After the signal chain reshapes the move distribution, the top three candidates are evaluated through a 1-ply subjective lookahead.
Each candidate move is pushed onto the board, the resulting position is read through the brain's evolved Perception module, and the position's ``feel'' is computed as the sum of Perception outputs.
The candidate with the highest feel score is selected.
This is not search---it is personality-dependent evaluation.
Different brains have different Perception weights, so the same position feels different to each brain.

\paragraph{Piece Preference.}
Each brain evolves six piece-type weights that bias candidate selection toward or away from moves involving specific piece types.
This gives the brain a direct way to express piece-level preference.

\subsection{Move Predictors (Cartridges)}

The base move distribution comes from a neural move predictor, which we term the \emph{cartridge}.
The brain does not replace this predictor---it reshapes its output distribution.
This separates chess knowledge (the cartridge) from chess personality (the brain), so the same brain architecture can be tested on predictors of different capabilities.

We use two cartridges: (1)~the AILED Engine v26.3.0, a 43M-parameter encoder-decoder transformer trained on 400K human games, achieving 21.3\% top-1 move prediction accuracy on Lichess games in the 1000--1800~Elo range (42.5\% top-3, 54.8\% top-5, measured on 5{,}976 positions).
Unlike conventional predictors, AILED Engine outputs win/draw/loss (WDL) estimates and piece-level attention weights alongside move probabilities---the first chess predictor designed to expose perceptual information to an external cognitive layer.
Its move prediction is weak by design: it was built for \emph{expressiveness}, not accuracy.
(2)~Maia2~\citep{mcilroyyoung2022maia2}, a pre-trained human-move predictor trained on millions of games and calibrated to specific Elo ratings (1100, 1400, 1600), which outputs move probabilities only but with substantially higher prediction accuracy.
The richer output of AILED Engine gives the brain's Perception, Affect, and Attention modules meaningful input signals, enabling imagination to differentiate across seeds.

\section{Evolution and Lifetime Learning}

The system operates on two timescales: \emph{genetic evolution} across generations and \emph{Hebbian learning} within each game.

\subsection{Genetic Evolution (NEAT)}

Each generation, the population of 20 brains is evaluated by playing 20 games against calibrated opponents.
The top 50\% survive; the population is filled by crossover (75\%) and cloned mutation (25\%).
The best genome is preserved unchanged via elitism.
Mutation operators follow standard NEAT~\citep{stanley2002neat}: weight perturbation (80\% of connections, $\mathcal{N}(0, 0.3)$), add node (3\%), add connection (5\%), and inter-module wiring perturbation (10\%).

Fitness in the ``multi'' mode used for all reported experiments is:
\begin{equation}
    F = 0.6 \cdot A + 0.2 \cdot C + 0.2 \cdot W
    \label{eq:fitness}
\end{equation}
where $A$ is move agreement with the cartridge, $C$ is a calibration score (agree when the cartridge is confident, explore when it is not), and $W$ is win rate (draws count as 0.5).
Each component is averaged over 20 games per evaluation.
Hebbian rewards are also per-module: Perception and Personality optimize agreement, Dynamics optimizes calibration, Affect rewards psyche--material correlation, Attention rewards agreement on confident positions, and Memory rewards improvement from the first half of a game to the second.

\subsection{Lifetime Learning (Hebbian Plasticity)}
\label{sec:hebbian}

Within each game, Hebbian learning adjusts connection weights based on correlated activations and game outcome.
After each move, every connection $c$ in all eight modules is updated:
\begin{equation}
    \Delta w_c = \eta \cdot a_{\text{pre}} \cdot (r - b)
    \label{eq:hebbian}
\end{equation}
where $\eta = 0.01$ is the base learning rate, $a_{\text{pre}}$ is the presynaptic activation, $r$ is the immediate reward signal, and $b$ is a running baseline.

Two mechanisms prevent destabilization: drift bounds ($\pm 0.3$ from genetic base) and anchor decay (weights pulled back toward base by factor $0.01$ after each update).
These constraints ensure that Hebbian modifications are temporary---they adapt behavior within a game but do not directly alter the genome passed to offspring.
This is what makes the Baldwin Effect possible: learned traits are not inherited, but genomes that learn faster reproduce more.

\subsection{Interaction Between Timescales}

The Baldwin Effect lives in the gap between these timescales: genomes that can learn within a lifetime survive and reproduce, even if their base weights are imperfect.
Over generations, mutations that approximate the learned behavior gain an edge.
Without Hebbian (OFF condition), the population depends on mutation alone.

\section{Experiments}

\subsection{Experimental Design}

We conduct experiments organized into four sets (Table~\ref{tab:conditions}).
Set~1 compares Hebbian ON versus OFF with 10~seeds each, using a heterogeneous matchup (different engines for cartridge and opponent).
Set~2 tests same-model matchups at various Elo gaps.
Set~3 replaces the same-model opponent with a different engine at the same skill level.
Set~4 tests fitness weight sensitivity.

\begin{table}[t]
\caption{Experimental conditions. All: population 20, desirability-domain signal chain, multi-objective fitness unless noted.}
\label{tab:conditions}
\centering
\small
\begin{tabular}{@{}llllr@{}}
\toprule
Experiment & Cartridge & Opponent & Hebb & Gens \\
\midrule
\multicolumn{5}{@{}l}{\emph{Set 1: Hebbian ON vs OFF (10 seeds each)}} \\
v26-ON & AILED Engine v26.3.0 & Maia2 1100 & ON & 50 \\
v26-OFF & AILED Engine v26.3.0 & Maia2 1100 & OFF & 50 \\
\midrule
\multicolumn{5}{@{}l}{\emph{Set 2: Same-model (mirror) matchups}} \\
Dominant & Maia2 1400 & Maia2 1100 & ON & 100 \\
Match-1400 & Maia2 1400 & Maia2 1400 & ON & 100 \\
Match-1600 & Maia2 1600 & Maia2 1600 & ON & 50 \\
\midrule
\multicolumn{5}{@{}l}{\emph{Set 3: Heterogeneous opponent (3 seeds)}} \\
Stockfish & Maia2 1600 & Stockfish $\sim$1600 & ON & 20 \\
\midrule
\multicolumn{5}{@{}l}{\emph{Set 4: Fitness weight sensitivity (4 seeds each)}} \\
Equal-ON & AILED Engine v26.3.0 & Maia2 1100 & ON & 50 \\
Equal-OFF & AILED Engine v26.3.0 & Maia2 1100 & OFF & 50 \\
\midrule
\multicolumn{5}{@{}l}{\emph{Set 5: Imagination ablation (3 seeds)}} \\
No-Imag & AILED Engine v26.3.0 & Maia2 1100 & ON & 50 \\
\bottomrule
\end{tabular}
\end{table}

\paragraph{Metrics.}
We report four metrics: \emph{fitness} (multi-objective score combining agreement, calibration, and win rate, scaled 0--1), \emph{agreement} (fraction of moves where the brain's selected move matches the cartridge's top-1 prediction), \emph{convergence generation} (first generation at which best fitness stabilizes within $\pm 0.01$ for 10+ generations), and \emph{late-stage slope} (linear slope of best fitness over the final 10 generations).

\paragraph{Baselines.}
Without any brain, the AILED Engine v26.3.0 cartridge agrees with Maia2-1100 on 22.5\% of moves and wins 0/30 games.
A random-parameter signal chain performs worse (19.6\% agreement, 3\% win rate)---random modulation disrupts the cartridge.
Temperature-only modulation peaks at 30.6\% agreement ($T{=}0.5$) or 10\% win rate ($T{=}3.0$), but no temperature achieves both.
Even Maia2 with temperature ($T{=}1.5$) drops from 50\% to 5\% win rate against its own argmax---noise hurts any model.
The evolved brain (40.4\% agreement, 42.5\% win rate) outperforms all non-evolutionary baselines by $>$30 percentage points in win rate.

\paragraph{Prior experiments.}
An earlier architecture (probability-domain signal chain, no imagination, Maia2 as both cartridge and opponent) showed ON/OFF separation over 150~generations but produced negligible cross-seed variance (0.05\% spread).
This motivated the current redesign with the desirability-domain signal chain, imagination, and AILED Engine's richer output.

\subsection{Results}

\subsubsection{Regime 1: Exploration (Hebbian ON)}

\begin{figure}[t]
\centering
\begin{tikzpicture}
\begin{groupplot}[
    group style={
        group size=2 by 1,
        horizontal sep=1.8cm,
    },
    width=0.48\textwidth,
    height=5cm,
    grid=major,
    grid style={gray!20},
    xlabel={Generation},
    legend style={font=\footnotesize, at={(0.02,0.98)}, anchor=north west},
    tick label style={font=\small},
    label style={font=\small},
]

\nextgroupplot[ylabel={Best Fitness}, ymin=0.30, ymax=0.55]
\addplot[blue!15, forget plot, no markers] coordinates {
    (1,0.3801) (6,0.3991) (11,0.4308) (16,0.4288) (21,0.4368)
    (26,0.4547) (31,0.4685) (36,0.4935) (41,0.5001) (46,0.5180) (50,0.5182)
} \closedcycle;
\addplot[blue!15, forget plot, no markers, fill=blue!10] coordinates {
    (1,0.3303) (6,0.3500) (11,0.3700) (16,0.3774) (21,0.3944)
    (26,0.4115) (31,0.4072) (36,0.4105) (41,0.4137) (46,0.4178) (50,0.4243)
};
\addplot[blue, thick, mark=none] coordinates {
    (1,0.3552) (6,0.3745) (11,0.4004) (16,0.4031) (21,0.4156)
    (26,0.4331) (31,0.4379) (36,0.4520) (41,0.4569) (46,0.4679) (50,0.4713)
};
\addlegendentry{ON ($n=10$)}
\addplot[red, thick, mark=none, dashed] coordinates {
    (1,0.3880) (6,0.4322) (11,0.4627) (16,0.4631) (21,0.4665)
    (26,0.4751) (31,0.4747) (36,0.4856) (41,0.4837) (46,0.4845) (50,0.4846)
};
\addlegendentry{OFF ($n=10$)}

\nextgroupplot[ylabel={Agreement (\%)}, ymin=20, ymax=55]
\addplot[blue, thick, mark=none] coordinates {
    (1,27.84) (6,31.13) (11,32.30) (16,32.79) (21,35.03)
    (26,37.10) (31,37.10) (36,38.78) (41,39.82) (46,40.38) (50,40.40)
};
\addlegendentry{ON ($n=10$)}
\addplot[red, thick, mark=none, dashed] coordinates {
    (1,32.61) (6,34.23) (11,38.09) (16,39.35) (21,40.36)
    (26,40.20) (31,40.67) (36,41.88) (41,41.62) (46,41.65) (50,41.80)
};
\addlegendentry{OFF ($n=10$)}

\end{groupplot}
\end{tikzpicture}
\caption{Hebbian ON vs OFF trajectories (mean across 10~seeds each) with AILED Engine v26.3.0 against Maia2 at 1100~Elo. Left: mean best fitness. OFF rises faster but plateaus by gen~20; ON climbs steadily through gen~50. Right: mean agreement. Both approach $\sim$40--42\%, but ON's late-stage slope is $7.6\times$ larger than OFF's. The difference is in dynamics, not endpoint.}
\label{fig:onoff}
\end{figure}
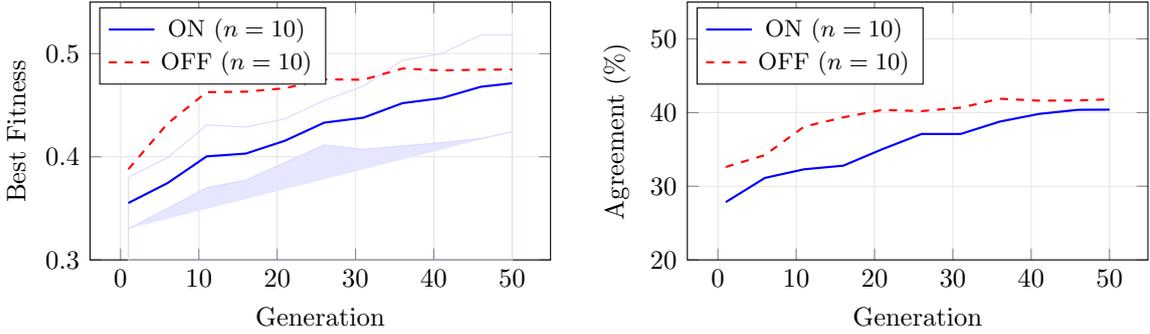

With the AILED Engine v26.3.0 cartridge playing against Maia2 at 1100~Elo, the Hebbian ON condition produces steadily climbing trajectories that have not plateaued at generation~50 (Figure~\ref{fig:onoff}, Table~\ref{tab:onoff}).
Across 10~seeds, mean best fitness reaches $0.471 \pm 0.047$ and mean agreement $40.4\% \pm 7.4\%$.
Four of ten ON seeds are still improving at experiment's end (late-stage slope $0.00140$, vs $0.00018$ for OFF---a $7.6\times$ ratio).

The defining property of this regime is a \emph{variance crossover}.
Early in evolution (generation~11), ON has \emph{lower} cross-seed agreement variance than OFF---Hebbian learning initially compresses diversity as all seeds learn similar early lessons.
By generation~34, ON surpasses OFF, and the variance ratio increases monotonically through generation~50 (Spearman $\rho = 0.91$, $p < 10^{-6}$; Figure~\ref{fig:onoff}).
At the endpoint, agreement ranges from 31.7\% (seed~9) to 54.1\% (seed~2)---a 22.4-point spread, $2.36\times$ higher variance than OFF (permutation test $p = 0.041$; bootstrap 95\% CI $[0.78, 6.35]$; Levene's test $p = 0.367$ at 14\% power with $n = 10$).
Each seed evolves different Perception weights; imagination amplifies those differences through 1-ply lookahead using AILED Engine's WDL and piece-focus signals; Hebbian learning reinforces each seed's emerging preferences during each game.
The result is that each seed evolves its own behavioral divergence---stochastic by evolution.

Win rates are also diverse: 0.17 to 0.65 across seeds.
ON brains take bigger risks, winning more in some seeds and losing more in others.
The population maintains internal diversity throughout: the best--average fitness gap is $0.068$ at generation~50, indicating the elite does not crush the population.

\subsubsection{Regime 2: Lottery Evolution (Hebbian OFF)}

The Hebbian OFF condition reaches similar mean fitness ($0.485 \pm 0.028$) and agreement ($41.8\% \pm 4.8\%$), but through a fundamentally different mechanism: discrete lucky jumps followed by elitism lock-in.

One seed illustrates the lottery clearly.
Seed~3 finds an elite genome that achieves 0.75 win rate and 0.542 fitness---the highest of any seed in either condition.
Eight of ten OFF seeds park at exactly 0.50 win rate, compared to only one of ten ON seeds (Fisher exact test, $p = 0.020$).
This is the lottery: most seeds draw the average ticket; a rare outlier hits the jackpot.

Elite lock-in is pervasive.
Seed~10 locks its best genome at generation~3 and holds it unchanged for 48~generations.
Seed~2 locks at generation~10 (41~gens frozen), seed~7 at generation~25 (26~gens frozen), seed~8 at generation~36 (15~gens frozen).
The late-stage slope is near zero ($0.00018$ mean across seeds), though one seed (off-seed6) remained active late.
Without Hebbian smoothing, the fitness landscape is too rugged for incremental improvement once elitism preserves a decent genome.

The cross-seed agreement spread is 14.1 points (33.9\%--48.0\%), narrower than ON's 22.4 points.
Lower variance is expected: without lifetime learning, evolution has fewer navigable paths, so most seeds converge to similar local optima.

\begin{table}[t]
\caption{Hebbian ON vs OFF with AILED Engine v26.3.0 cartridge against Maia2 at 1100~Elo. 50 generations, population 20, 20 games/eval. 10~seeds per condition.}
\label{tab:onoff}
\centering
\small
\begin{tabular}{@{}lrr@{}}
\toprule
Metric & Hebbian ON & Hebbian OFF \\
\midrule
Seeds ($n$) & 10 & 10 \\
Mean fitness & $0.471 \pm 0.047$ & $0.485 \pm 0.028$ \\
Mean agreement & $40.4\% \pm 7.4\%$ & $41.8\% \pm 4.8\%$ \\
Agreement range & 31.7\%--54.1\% & 33.9\%--48.0\% \\
Mean win rate & $0.42 \pm 0.16$ & $0.49 \pm 0.12$ \\
Seeds at win $=$ 0.50 & 1 / 10 & 8 / 10 \\
\midrule
\multicolumn{3}{@{}l}{\emph{Endpoint comparisons (n.s.)}} \\
Welch's $t$ (fitness) & \multicolumn{2}{c}{$t = -0.77$, $p = 0.45$} \\
Welch's $t$ (agreement) & \multicolumn{2}{c}{$t = -0.50$, $p = 0.62$} \\
\midrule
\multicolumn{3}{@{}l}{\emph{Dynamics (significant)}} \\
Agreement variance & 0.0055 & 0.0023 \\
Variance ratio (ON/OFF) & \multicolumn{2}{c}{2.36$\times$ (perm.\ $p = 0.041$)} \\
Variance crossover & \multicolumn{2}{c}{$\rho = 0.91$, $p < 10^{-6}$} \\
Win-rate clustering & \multicolumn{2}{c}{Fisher exact $p = 0.020$} \\
Late-stage slope ($\times 10^{-4}$) & 14.0 & 1.8 \\
Slope ratio & \multicolumn{2}{c}{7.6$\times$ (perm.\ $p = 0.040$, $d = 0.76$)} \\
Converged by gen~50 & 4 / 10 & 7 / 10 \\
ICC(1) (last 20 gens) & 0.81 & 0.92 \\
\bottomrule
\end{tabular}
\end{table}

\subsubsection{Regime 3: Brain Transparency (Same-Model Opponents)}

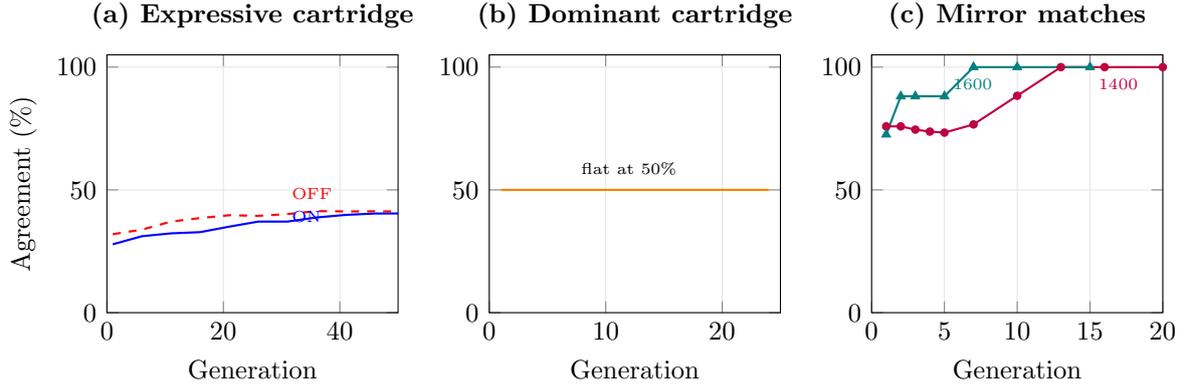
\begin{figure}[t]
\centering
\begin{tikzpicture}
\begin{groupplot}[
    group style={
        group size=3 by 1,
        horizontal sep=1.2cm,
    },
    width=0.34\textwidth,
    height=5cm,
    grid=major,
    grid style={gray!20},
    xlabel={Generation},
    ylabel={Agreement (\%)},
    tick label style={font=\small},
    label style={font=\small},
    title style={font=\small\bfseries},
    ymin=0, ymax=105,
]

\nextgroupplot[title={(a) Expressive cartridge}, xmin=0, xmax=50]
\addplot[blue, thick, mark=none] coordinates {
    (1,27.84) (6,31.13) (11,32.30) (16,32.79) (21,35.03)
    (26,37.10) (31,37.10) (36,38.78) (41,39.82) (46,40.38) (50,40.40)
};
\addplot[red, thick, mark=none, dashed] coordinates {
    (1,31.96) (6,33.81) (11,37.09) (16,38.54) (21,39.68)
    (26,39.44) (31,40.15) (36,41.37) (41,41.23) (46,41.29) (50,41.20)
};
\node[font=\tiny, blue, anchor=south west] at (axis cs:30,33) {ON};
\node[font=\tiny, red, anchor=south west] at (axis cs:30,42) {OFF};

\nextgroupplot[title={(b) Dominant cartridge}, xmin=0, xmax=25, ylabel={}]
\addplot[orange, thick, mark=none] coordinates {
    (1,50.0) (5,50.0) (10,50.0) (15,50.0) (20,50.0) (24,50.0)
};
\node[font=\tiny, anchor=south] at (axis cs:12,52) {flat at 50\%};

\nextgroupplot[title={(c) Mirror matches}, xmin=0, xmax=20, ylabel={}]
\addplot[purple, thick, mark=*, mark size=1.2pt] coordinates {
    (1,75.87) (2,75.87) (3,74.55) (4,73.75) (5,73.33)
    (7,76.67) (10,88.33) (13,100) (16,100) (20,100)
};
\addplot[teal, thick, mark=triangle*, mark size=1.5pt] coordinates {
    (1,72.57) (2,88.12) (3,88.12) (5,88.12) (7,100) (10,100) (15,100)
};
\node[font=\tiny, purple, anchor=east] at (axis cs:19,93) {1400};
\node[font=\tiny, teal, anchor=east] at (axis cs:9,93) {1600};

\end{groupplot}
\end{tikzpicture}
\caption{Agreement trajectories across the three regimes. (a)~Expressive cartridge (AILED Engine v26.3.0 vs Maia2-1100, mean of 10~seeds per condition): both conditions approach $\sim$40\% agreement, but ON is still climbing while OFF has flatlined. The difference is in dynamics, not endpoint. (b)~Dominant cartridge (Maia2 1400 vs 1100, first 25 of 100 generations shown): agreement flat at 50\% throughout---the brain is passive. (c)~Mirror matches: agreement converges to 100\% (self-erasure). Maia2-1600 reaches transparency at gen~7, faster than 1400 at gen~13.}
\label{fig:regimes}
\end{figure}

Four experiments test what happens when cartridge and opponent are the same model at various Elo levels (Table~\ref{tab:cartridge}, Figure~\ref{fig:regimes}).
In every case, the brain converges toward transparency---regardless of the skill gap between cartridge and opponent.

\paragraph{Dominant cartridge (Maia2 1400 vs 1100).}
With a 300-Elo advantage, the cartridge wins without help.
Best fitness reaches 0.6000 at generation 1 and never changes.
Agreement sits at 50.0\%---the pass-through baseline.

\paragraph{Mirror matches (1400 vs 1400, 1600 vs 1600).}
When cartridge and opponent are identical, the brain rapidly learns perfect transparency.
At 1400 Elo, agreement climbs from 75.9\% to 100\% by generation~13.
At 1600 Elo, the same happens by generation~7---the cleaner predictor means less noise for the imagination loop to close.
The optimal strategy is to not interfere: any signal chain modulation can only hurt against your own twin.

\paragraph{Underdog (Maia2 1400 vs 1600).}
Even when the cartridge is \emph{weaker} than the opponent, the brain still erases itself---100\% agreement by generation~8.
The brain cannot improve on the cartridge by modulating it against a stronger version of the same model.
This rules out skill gap as the driver: the brain goes transparent whenever cartridge and opponent share the same error patterns.

\paragraph{Heterogeneous opponent (Maia2 1600 vs Stockfish $\sim$1600).}
When the opponent is a different engine at the same skill level, the pattern reverses.
Across 3~seeds (20~generations each), agreement stabilizes at $30.7\% \pm 1.3\%$ (range 29.3\%--31.7\%)---never trending toward 100\%.
The tight cross-seed variance ($\pm 1.3\%$) contrasts with the wide spread in the exploration regime ($\pm 7.4\%$): against a heterogeneous opponent at matched Elo, the brain finds a consistent personality rather than diverse strategies.
This confirms that the transparent regime is a mirror artifact, not a property of strong predictors.

\begin{table}[t]
\caption{Same-model and heterogeneous opponent experiments (all Hebbian ON).}
\label{tab:cartridge}
\centering
\small
\begin{tabular}{@{}llrr@{}}
\toprule
Experiment & Opponent type & Agree & Gens \\
\midrule
1400 vs 1100 & Same (dominant) & 50\%$^*$ & 100 \\
1400 vs 1400 & Same (mirror) & 100\% at g13 & 100 \\
1600 vs 1600 & Same (mirror) & 100\% at g7 & 50 \\
1400 vs 1600 & Same (underdog) & 100\% at g8 & 20 \\
\midrule
1600 vs SF 1600 & \textbf{Different (3 seeds)} & \textbf{30.7\% $\pm$ 1.3\%} & 20 \\
\bottomrule
\multicolumn{4}{@{}l}{\footnotesize $^*$Flat at 50\% throughout; brain contributes nothing.}
\end{tabular}
\end{table}

\subsubsection{The Agreement Ceiling}

The AILED Engine v26.3.0 cartridge's raw top-1 agreement with Maia2-1100 is 22.5\% (null-brain baseline, no signal chain).
Evolved brains raise this to a mean of $\sim$40\%, with one seed reaching 54.1\%---more than doubling the raw agreement.
In mirror-match experiments with Maia2, where the cartridge \emph{is} the reference, agreement reaches 100\% because perfect alignment with the cartridge is perfect alignment with the opponent's predictor.
There is no gap because there is no personality---the brain has learned to be invisible.

With the AILED Engine cartridge against a heterogeneous opponent, agreement across 10~ON seeds ranges from 31.7\% to 54.1\%.
The wide range reflects different evolved Perception weights producing different subjective evaluations via imagination---personality expressed as variance across seeds rather than a single gap below a ceiling.

\subsubsection{Hebbian Ablation on Evolved Genomes}

To confirm that evolved ON genomes depend on lifetime learning, we evaluated two brains from the exploration regime---the highest-agreement genome (seed~2, ``diplomat'': 54.1\% agreement, 45\% win rate during evolution) and the highest-win-rate genome (seed~9, ``fighter'': 31.7\% agreement, 65\% win rate during evolution)---with Hebbian disabled, then re-enabled.
Each brain played 100 games against Maia2 at three Elo levels (Table~\ref{tab:ablation}).

\begin{table}[t]
\caption{Hebbian ablation on two evolved ON genomes vs Maia2 at three Elo levels (100 games each). Same genome; only lifetime learning differs.}
\label{tab:ablation}
\centering
\small
\begin{tabular}{@{}llrrrrrr@{}}
\toprule
 & & \multicolumn{2}{c}{vs 1100} & \multicolumn{2}{c}{vs 1400} & \multicolumn{2}{c}{vs 1600} \\
Brain & Hebb & Win & Agr & Win & Agr & Win & Agr \\
\midrule
Diplomat & OFF & 25\% & 44\% & 32\% & 46\% & 25\% & 48\% \\
         & ON  & \textbf{41\%} & \textbf{52\%} & \textbf{44\%} & \textbf{52\%} & \textbf{37\%} & \textbf{48\%} \\
\midrule
Fighter  & OFF & 34\% & 20\% & 25\% & 25\% & 17\% & 20\% \\
         & ON  & 25\% & 21\% & 4\% & 19\% & 5\% & 22\% \\
\bottomrule
\end{tabular}
\end{table}

The diplomat genome shows the expected pattern: Hebbian ON raises win rate by 10--16 percentage points and agreement by 4--8 points across all opponents.
The genome encodes learnability, not competence---half its performance comes from lifetime learning.
Without Hebbian, the base weights still outperform the null baseline (22.5\% agreement, 44.1\% vs 22.5\%), demonstrating partial \emph{genetic assimilation}: traits originally requiring lifetime learning have been partially absorbed into the evolved base weights.
This is the classical signature of the Baldwin Effect---learned behaviors reshaping the fitness landscape until genetic variants that approximate the learned behavior are selected~\citep{baldwin1896}.

The fighter genome reveals a different strategy.
Its agreement (19.8--25.1\%) sits \emph{below} the null baseline---the signal chain actively steers away from the opponent's expected moves.
With Hebbian ON, the fighter wins 17 games against Maia2-1100 (the only brain to win any games outright) but collapses against stronger opponents (3.5\% and 4.5\% win rate at 1400 and 1600).
This is an anti-agreement strategy: disrupt the opponent by playing unexpected moves, a tactic that works only when the opponent is weak enough to be disrupted.

The structural difference is not just in agreement levels but in how each brain reshapes the move distribution.
Across multiple positions, the diplomat evolves high temperature ($+0.25$, commits to the cartridge's top choice), high saturation ceiling ($0.91$, lets favorites dominate), and low exploration ($0.04$).
The fighter evolves the opposite: negative temperature ($-0.45$, randomizes), low saturation ($0.23$, flattens the distribution), and high exploration ($0.20$).
Their EQ profiles are inverted: the diplomat boosts second-tier moves (band~2 gain $2.78$) while the fighter boosts the extremes (band~1 gain $2.07$, band~5 gain $1.98$).
Piece preferences also differ: the diplomat favors pawns ($2.78$) while the fighter favors queens ($1.46$).

These structural differences manifest as different chess, not different sampling.
On 8~diverse test positions (opening through endgame), the diplomat and fighter select \emph{different moves} 62\% of the time when given the same board and the same cartridge output.
Both also diverge from the cartridge's argmax: in the starting position, the cartridge plays 1.e4, the diplomat plays 1.Nf3, and the fighter plays 1.c4---three different opening choices from the same underlying predictor.
The diplomat consistently redirects toward knight development (Nf3 in 4/8 positions), while the fighter selects moves that neither the diplomat nor the cartridge would choose.

Over 30~games each, these move-level differences compound into distinct play styles.
The diplomat plays short games (27 moves average) with 30\% knight usage and never deploys queen or rook.
The fighter plays longer games (40 moves average) with diverse piece usage including queen (6\%) and rook (4\%).
Different openings, different piece repertoires, different game lengths---behavioral divergence that is stable, interpretable, and visible at every level from individual moves to full games.

Both brains evolved under the same fitness function, in the same regime, with the same architecture.
The diplomat and fighter represent two points on the 22.4-point agreement spread---different evolutionary paths to competence with structurally distinct signal chain configurations, distinct opening choices, and distinct piece-usage profiles.
The diplomat learns to align; the fighter learns to diverge.

\begin{figure}[t]
\centering
\begin{tikzpicture}
\begin{groupplot}[
    group style={
        group size=1 by 2,
        vertical sep=0.8cm,
    },
    width=\columnwidth,
    height=3.5cm,
    grid=major,
    grid style={gray!20},
    xlabel={Generation},
    tick label style={font=\small},
    label style={font=\small},
    legend style={font=\footnotesize, at={(0.98,0.98)}, anchor=north east},
    xmin=5, xmax=55,
]
\nextgroupplot[ylabel={Saturation}, ymin=0.1, ymax=1.0]
\addplot[blue, thick, mark=square*, mark size=1.5pt] coordinates {
    (10,0.4618) (20,0.5127) (30,0.8530) (40,0.9148) (50,0.9143)
};
\addlegendentry{Diplomat}
\addplot[red, thick, mark=triangle*, mark size=1.5pt, dashed] coordinates {
    (10,0.6295) (20,0.7539) (30,0.5704) (40,0.3683) (50,0.2230)
};
\addlegendentry{Fighter}

\nextgroupplot[ylabel={Temperature}, ymin=-0.6, ymax=0.5]
\addplot[blue, thick, mark=square*, mark size=1.5pt] coordinates {
    (10,-0.0462) (20,-0.2839) (30,0.3760) (40,0.3579) (50,0.2565)
};
\addplot[red, thick, mark=triangle*, mark size=1.5pt, dashed] coordinates {
    (10,0.2190) (20,-0.2888) (30,-0.0618) (40,-0.1206) (50,-0.4430)
};
\end{groupplot}
\end{tikzpicture}
\caption{Parameter evolution of the diplomat (seed~2) and fighter (seed~9) across generations 10--50. Top: saturation ceiling diverges from similar starting values to opposite extremes (0.91 vs 0.22). Bottom: temperature modifier diverges from near-zero to opposite signs ($+0.26$ vs $-0.44$). Divergence is learned progressively, not present at initialization.}
\label{fig:param_evolution}
\end{figure}
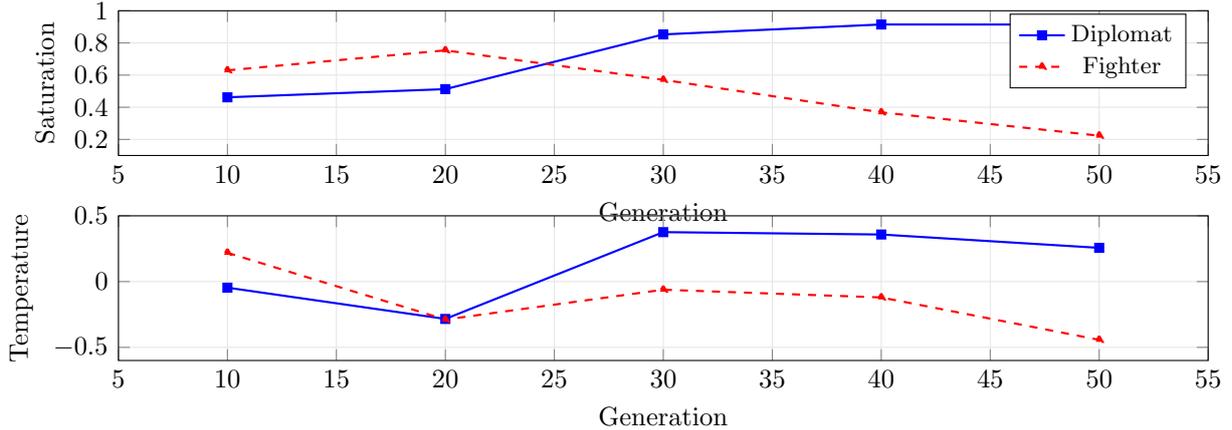

\subsubsection{Fitness Weight Sensitivity}

To test robustness, we repeated the ON/OFF comparison with equal weights ($F = (A + C + W) / 3$) instead of agreement-dominated weighting ($0.6/0.2/0.2$), using 4~seeds per condition.
The regime structure survives: ON agreement spread remains wider than OFF (24.4 vs 17.0 points; variance ratio $1.68\times$, same direction as multi-weight's $2.36\times$).
Equal weights enable extreme strategies: one ON seed evolves 10.4\% agreement with 75\% win rate, disagreeing with the cartridge on 90\% of moves.
OFF seeds cluster at 50\% win rate under both weightings (7/10 multi, 3/4 equal).
The fitness weights determine \emph{which} strategies evolve, not \emph{whether} ON produces more diverse organisms than OFF.

\subsubsection{Imagination Ablation}

To isolate imagination's role in the causal chain (Hebbian $\to$ Perception differences $\to$ imagination $\to$ amplified diversity), we ran the Hebbian ON condition with imagination disabled: the signal chain reshapes the distribution but the 1-ply Perception lookahead is removed.
Three seeds, same setup as Set~1 (AILED Engine v26.3.0 cartridge, Maia2-1100 opponent, 50~generations).

Without imagination, mean fitness drops to $0.334 \pm 0.005$ (vs $0.471$ with imagination), agreement to $29.3\% \pm 2.8\%$ (vs $40.4\%$), and win rate to $10.8\% \pm 5.1\%$ (vs $42.5\%$).
Agreement barely exceeds the null baseline (22.5\%), and the tight cross-seed variance ($\pm 2.8\%$ vs $\pm 7.4\%$) confirms that imagination is the diversity amplifier: without it, different Perception weights produce negligible behavioral divergence.
The effect is large and unambiguous despite $n = 3$: the 95\% CI for agreement without imagination is approximately $[25.0\%, 33.6\%]$, entirely below the with-imagination mean of $40.4\%$.
A Welch's $t$-test yields $t = 3.47$, $p = 0.004$ (Cohen's $d = 1.89$).
No seed without imagination reaches even the lowest seed with imagination.
This completes the causal chain: Hebbian creates Perception differences (ON/OFF comparison, $n = 10$, $p = 0.041$), imagination amplifies them into behavioral divergence (this ablation, $n = 3$, $p = 0.004$), and the variance crossover ($\rho = 0.91$, $p < 10^{-6}$) measures the combined effect.

\section{Discussion}

\subsection{Three Regimes: Baldwin-Compatible Dynamics}

Turning Hebbian learning on or off does not just speed up or slow down a single process.
It produces dynamics compatible with the Baldwin Effect, and the switch is visible as a variance crossover.

Early in evolution, ON has \emph{lower} cross-seed variance than OFF.
Hebbian learning initially compresses diversity: all seeds learn similar early lessons (higher agreement with the opponent, better calibration), so they cluster together.
This is consistent with Paenke et al.'s~\citep{paenke2009baldwin} prediction that plasticity reduces variance by buffering organisms against environmental noise.
But this compression is temporary.
By generation~34, ON surpasses OFF in variance, and the ratio increases monotonically through generation~50 (Spearman $\rho = 0.91$, $p < 10^{-6}$).
The reversal occurs because Hebbian learning does not just buffer---it \emph{amplifies} seed-specific Perception differences through the imagination feedback loop, producing divergent behavioral strategies that mutation alone cannot reach.

The \emph{exploration} regime is thus a two-phase process: initial compression (Paenke's prediction holds), then expansion as imagination and Hebbian plasticity interact to open new regions of the fitness landscape.
The endpoint variance ratio ($2.36\times$, permutation $p = 0.041$) is a snapshot of an ongoing trend, not a static property.

Crucially, the cross-seed variance reflects genuine behavioral differences, not noise.
The intraclass correlation coefficient across the last 20~generations is ICC $= 0.81$ for ON and $0.92$ for OFF (both $> 0.75$, indicating strong consistency).
Each seed maintains a distinct, stable agreement level across generations---the diplomat (seed~2, $\sim$50\% agreement) and fighter (seed~9, $\sim$33\%) represent reproducible behavioral signatures, not random fluctuations.
The Hebbian ablation (Table~\ref{tab:ablation}) confirms this structurally: the diplomat uses Hebbian for alignment while the fighter uses it for disruption---qualitatively different strategies, not different draws from the same distribution.

The \emph{lottery} regime (Hebbian OFF) reaches similar mean endpoints but through discrete lucky mutations followed by elitism lock-in.
Eight of ten OFF seeds park at exactly 0.50 win rate (vs one of ten ON seeds; Fisher exact $p = 0.020$).
Without Hebbian smoothing, the landscape is too rugged for incremental improvement: once a decent genome locks, the population stops.
The late-stage slope is effectively zero (permutation $p = 0.040$ vs ON, Cohen's $d = 0.76$).
With 10~seeds, the lottery is visible at scale: most tickets are mediocre, rare outliers peak high.

The \emph{transparent} regime is a robust empirical phenomenon.
The pattern is consistent across every tested condition: same-model opponents produce self-erasure (100\% agreement in 7--13 generations) regardless of skill gap, while a different engine at the same Elo does not (30.7\% $\pm$ 1.3\% across 3~seeds).

In every condition tested, the brain converges to transparency when cartridge and opponent share a model architecture, and retains behavioral divergence when they do not.
The evidence supporting ``if'': all four same-model conditions (dominant, mirror-1400, mirror-1600, underdog) produce self-erasure, despite varying Elo gap from $-200$ to $+300$.
The evidence supporting ``only if'': the Stockfish control at matched Elo ($\sim$1600) produces 30.7\% agreement with no trend toward 100\%.

What remains unresolved is the mechanism.
The simplest explanation---same architecture produces correlated errors, leaving no gradient for the signal chain to exploit---is consistent with all observations but not uniquely supported.
Alternative explanations include shared move-ranking geometry (same relative ordering even at different Elo) or shared positional blindspots from shared training data.
Disentangling these would require opponents that share architecture but differ in training data, or vice versa---conditions not available with current models.
We present the transparent regime as an empirical regularity with a plausible but not proven mechanism, and note that it generates a concrete, testable prediction for any system using self-play.

All three regimes come from the same architecture, the same NEAT operators, and the same Hebbian rule.
What changes is the relationship between cartridge and opponent.

\subsection{Personality Requires Heterogeneous Competition}

Cognitive personality---the brain's systematic disagreement with the cartridge---only shows up when the opponent plays differently from the cartridge.
With 10~seeds per condition, personality manifests not as a single gap but as variance: ON seeds spread across 22.4~points of agreement, each evolving its own subjective evaluation.

Four observations support this.
First, the AILED Engine cartridge playing against Maia2 (a different engine) produces brains with diverse, non-trivial preferences across all 20~seeds (ON and OFF).
Second, all same-model Maia2 matchups produce brains that erase themselves, regardless of Elo gap (dominant, matched, or underdog).
Third, convergence speed to transparency scales with predictor quality in same-model matches (gen~13 at 1400, gen~7 at 1600).
Fourth, replacing the Maia2 opponent with Stockfish at the same Elo blocks the self-erasure entirely---agreement stays at $\sim$30\%.

The brain adds value when it can exploit the mismatch between how the cartridge plays and how the opponent plays.
When there is no mismatch, the brain is overhead.

\subsection{The Brain as Competence Amplifier}

The AILED Engine cartridge alone (21.3\% top-1 accuracy, 0/30 wins) cannot compete against Maia2-1100.
With an evolved brain, it reaches $\sim$40\% agreement and $\sim$43--52\% win rate.
The brain does not learn chess---it learns where to disagree with its own cartridge, re-ranking the cartridge's top-5 candidates (54.8\% of which include the opponent's preferred move) via imagination and evolved Perception.
Each seed discovers its own path to competence, producing the 22.4-point agreement spread.

\subsection{Imagination: Deterministic to Stochastic by Evolution}

Imagination previews candidate moves through evolved Perception.
With Maia2 (probability-only output), cross-seed spread was 0.05\%---imagination collapsed to near-deterministic evaluation.
With AILED Engine (WDL + piece attention), seeds evolve different Perception weights that respond differently to richer signals, producing a 22.4-point spread.
Hebbian amplifies this: within each game, weight updates reinforce each seed's emerging preferences.
The variance crossover ($\rho = 0.91$) measures the combined effect---evolved Perception diversity, amplified by imagination, reinforced by Hebbian plasticity.
In mirror matches, imagination finds no moves the cartridge missed and the loop closes to 100\% agreement; against heterogeneous opponents, WDL swings keep the loop open.

\subsection{The Agreement Ceiling as Cartridge Fingerprint}

Agreement is bounded by cross-engine move overlap---how often the cartridge's top-1 happens to match the opponent's.
The AILED Engine null baseline is 22.5\%; evolved brains raise this to $\sim$40\% mean (one seed reaches 54.1\%), more than doubling the raw overlap despite the two engines being trained on different data at different scales.
With 10~seeds, the ``personality gap'' becomes a distribution: each seed's distance from the ceiling reflects its evolved preferences, and this variance (ICC $= 0.81$) is itself a signature of behavioral divergence at the population level.

\subsection{Implications Beyond Chess}

Four findings generalize.
\emph{Self-play suppresses diversity}: same-model opponents produce self-erasing agents in 7--13 generations, suggesting that AlphaZero-style self-play~\citep{silver2018alphazero} may systematically suppress behavioral variety.
\emph{Expressiveness over accuracy}: a 21.3\%-accurate predictor with WDL and piece attention outperforms temperature-only modulation by $>$30 points in win rate when paired with an evolved cognitive layer---rich output matters more than raw prediction quality.
\emph{Plasticity's effect reverses over time}: variance compression (Paenke et al.'s prediction~\citep{paenke2009baldwin}) holds early but reverses after generation~34; any system evolving plastic networks should expect timescale-dependent effects.
\emph{Diversity is environmental}: the diplomat (1.Nf3, short games) and fighter (1.c4, long games) emerged from identical architecture and fitness but required a heterogeneous opponent to differentiate.

\subsection{Limitations}

Several limitations qualify these findings:

\begin{enumerate}
    \item \textbf{Generation count.} 50~generations characterize the \emph{trajectory} of the variance crossover, not its endpoint. OFF leads at the stopping point, and we cannot rule out that ON is simply slower. Our contribution is the trend reversal itself ($\rho = 0.91$), not the final magnitude. Longer runs (200+ generations) would clarify whether the crossover continues, saturates, or reverses again.

    \item \textbf{Single expressive cartridge.} Only one expressive cartridge (AILED Engine v26.3.0) was tested. The regime structure may depend on the specific richness of this cartridge's output. Testing with other multi-output predictors would strengthen generalizability.

    \item \textbf{Sample sizes.} The main comparison uses $n = 10$ seeds per condition; the equal-weight sensitivity analysis uses $n = 4$; the Stockfish heterogeneous control uses $n = 3$. Levene's test for variance equality has only 14\% power at $n = 10$, meaning the $2.36\times$ variance ratio could be a false positive at this sample size. We report the permutation test ($p = 0.041$) and bootstrap CI as more appropriate tests for small $n$, but acknowledge that $n = 20$+ would provide more robust conclusions.

    \item \textbf{Unidirectional interface.} The brain--cartridge interface is unidirectional---the brain reshapes the cartridge's output but the cartridge does not condition on brain state. Bidirectional co-adaptation could change the regime picture.

    \item \textbf{Imposed modularity and component isolation.} The modular structure (eight named modules) is imposed rather than evolved. We cannot claim that NEAT would discover modular organization on its own. Furthermore, of the system's interacting components (modules, signal chain, equalization, imagination, Hebbian plasticity), only imagination and Hebbian plasticity are ablated individually. The variance crossover could be driven by any combination of unablated components. A full factorial ablation would require matching parameter counts, input/output dimensionality, and training conditions across all combinations, which constitutes a separate study.

    \item \textbf{No human evaluation.} No Elo evaluation against human players. The behavioral divergence we measure is between agents; whether humans perceive these agents as having different ``personalities'' remains untested.

    \item \textbf{Imagination ablation uses $n = 3$.} The imagination ablation (Section~5.2) confirms imagination's role as the diversity amplifier, but uses only 3~seeds. The effect is large (agreement drops from 40.4\% to 29.3\%, win rate from 42.5\% to 10.8\%) with tight cross-seed variance ($\pm 2.8\%$), but $n = 10$ would strengthen the comparison.
\end{enumerate}

\section{Conclusion}

The central finding of this work is that \textbf{plasticity's effect on behavioral variance reverses over evolutionary time}: compression early, expansion later, with a monotonic crossover trend ($\rho = 0.91$, $p < 10^{-6}$).
Prior theory~\citep{paenke2009baldwin} predicts variance reduction; we observe this initially, but the trend reverses at generation~34 as imagination amplifies evolved Perception differences into divergent behavioral strategies that mutation alone cannot sustain.

This diversity is not noise.
Evolved agents select different moves on the same board (62\% disagreement between two seeds), develop distinct opening choices (1.Nf3 vs 1.c4), piece preferences (knight-focused vs queen-active), and game lengths (27 vs 40 moves).
Their signal chain configurations are structurally inverted across every major parameter.
These are reproducible behavioral signatures (ICC $> 0.8$), not sampling artifacts---and no non-evolutionary baseline (temperature tuning, random parameters) comes within 30 percentage points of the evolved brain's performance.

Three regimes characterize the interaction between plasticity and competition.
The exploration regime (Hebbian ON) produces this diversity.
The lottery regime (Hebbian OFF) reaches similar endpoints but through lucky mutations and elitism lock-in (Fisher exact $p = 0.020$).
The transparent regime (same-model opponent) erases all diversity---a falsifiable prediction that self-play systems may systematically suppress behavioral differentiation.

These results carry beyond chess.
Plasticity's timescale-dependent effect on variance applies to any system evolving plastic neural networks.
The transparent regime applies to any system using self-play.
And the finding that a weak predictor (21.3\% top-1 accuracy) becomes competitive through evolved cognitive modulation challenges the assumption that model accuracy is the primary bottleneck---expressiveness may matter more.

Three directions follow.
First, a bidirectional brain--cartridge interface where the predictor conditions on brain state, enabling co-adaptation.
Second, systematic variation of opponent similarity to map the boundary of the transparent regime.
Third, deploying evolved brains on Lichess to test whether behavioral divergence translates into measurable play-style differences against real humans.

\section*{Acknowledgments}

The author thanks the open-source communities behind python-chess, PyTorch,
and the Maia chess project for enabling this work.
Portions of this manuscript were drafted with the assistance of Claude
(Anthropic), an AI language model.

\paragraph{Reproducibility.}
Source code, trained model checkpoints (including the AILED Engine v26.3.0 predictor), experiment data, and analysis scripts will be available at \url{https://github.com/Ailed-AI/ailed-brain} upon publication.

\bibliographystyle{plainnat}
\bibliography{references}

\end{document}